\documentclass[conference, a4paper]{IEEEtran}
\IEEEoverridecommandlockouts
\usepackage{cite}
\usepackage{amsmath,amssymb,amsfonts}
\usepackage{algorithmic}
\usepackage{graphicx}
\usepackage[caption=false, font=footnotesize]{subfig}
\usepackage{textcomp}
\usepackage{xcolor}
\def\BibTeX{{\rm B\kern-.05em{\sc i\kern-.025em b}\kern-.08em
    T\kern-.1667em\lower.7ex\hbox{E}\kern-.125emX}}
\usepackage{eso-pic}
\usepackage{hyperref}

\AddToShipoutPictureBG*{%
	\AtPageUpperLeft{%
		\put(60,-30){%
			\parbox{0.8\paperwidth}{\centering
				\footnotesize
				This paper has been accepted for presentation at the 2023 IEEE Underwater Technology (UT) conference.\\
				Please cite the paper as: J. Su, X. Tu, F. Qu, et al. "Information-preserved blending method for forward-looking sonar mosaicing in non-ideal system configuration" 2023 IEEE Underwater Technology (UT).\\
				DOI: \href{https://doi.org/10.1109/UT49729.2023.10103367}{10.1109/UT49729.2023.10103367}\\
				URL: \href{https://ieeexplore.ieee.org/abstract/document/10103367}{https://ieeexplore.ieee.org/abstract/document/10103367}
			}
		}
	}
}

\begin{document}

\title{
    Information-Preserved Blending Method for Forward-Looking Sonar Mosaicing in Non-Ideal System Configuration
% \thanks{Identify applicable funding agency here. If none, delete this.}
}

\author{
    \IEEEauthorblockN{Jiayi Su*\thanks{* Jiayi Su is the corresponding author.}\IEEEauthorrefmark{2}, Xingbin Tu\IEEEauthorrefmark{2}, Fengzhong Qu\IEEEauthorrefmark{2}\IEEEauthorrefmark{3}, Yan Wei\IEEEauthorrefmark{2}}
    \IEEEauthorblockA{\IEEEauthorrefmark{2}\textit{Key Laboratory of Ocean Observation-Imaging Testbed of Zhejiang Province, Zhejiang University} \\
    \IEEEauthorrefmark{2}The Engineering Research Center of Oceanic Sensing Technology and Equipment, Ministry of Education \\
    	\textit{Zhoushan, China} \\
    \IEEEauthorblockA{\IEEEauthorrefmark{3}Hainan Institute of Zhejiang University \\
	   	\textit{Sanya, China} \\
    \textit{\{sujyzju, xbtu, jimqufz, redwine447\}@zju.edu.cn}
}
}
}

\maketitle

\begin{abstract}
    Forward-Looking Sonar (FLS) has started to gain attention in the field of near-bottom close-range underwater inspection because of its high resolution and high framerate features. Although Automatic Target Recognition (ATR) algorithms have been applied tentatively for object-searching tasks, human supervision is still indispensable, especially when involving critical areas. A clear FLS mosaic containing all suspicious information is in demand to help experts deal with tremendous perception data. However, previous work only considered that FLS is working in an ideal system configuration, which assumes an appropriate sonar imaging setup and the availability of accurate positioning data. Without those promises, the intra-frame and inter-frame artifacts will appear and degrade the quality of the final mosaic by making the information of interest invisible. In this paper, we propose a novel blending method for FLS mosaicing which can preserve interested information. A Long-Short Time Sliding Window (LST-SW) is designed to rectify the local statistics of raw sonar images. The statistics are then utilized to construct a Global Variance Map (GVM). The GVM helps to emphasize the useful information contained in images in the blending phase by classifying the informative and featureless pixels, thereby enhancing the quality of final mosaic. The method is verified using data collected in the real environment. The results show that our method can preserve more details in FLS mosaics for human inspection purposes in practice.
  \end{abstract}

\begin{IEEEkeywords}
    Forward-looking sonar, image mosaic, image blending, underwater inspection.
\end{IEEEkeywords}

\section{Introduction}
Forward-Looking Sonar (FLS) has received special attention because of its realistic imaging quality. It shows power in the field of underwater inspection, such as ship hull inspection \cite{hover2012advanced} and environment mapping. Forward-looking sonar mosaics present underwater information in a more compact pattern. It can significantly accelerate human inspection during underwater inspection tasks and reduce the storage requirements for data, provided that the quality of mosaics is guaranteed. However, due to the complex imaging model of FLS, this can not always be promised.

%\begin{figure}[tb]
%	\centerline{\includegraphics[width=3.5in]{figures/sonar_sketch.eps}}
%	\caption{Illustration of the imaging model of forward-looking sonar. $\theta$ is the bearing angle, $\phi$ is the elevation, $r$ is the range. The forward-looking sonar samples in time along each beam direction to obtain acoustic intensity data at different distances.}
%	\label{sonar_sketch}
%\end{figure}

The quality of mosaics can be either superior to the original representation \cite{hurtos2013novel}, or degenerated due to the presence of intra- and inter-frame artifacts. In the case of degeneration, the desired information even disappears. In this paper, we take both intra- and inter-frame artifacts into account and propose a Global Variance Map (GVM) based blending method for FLS mosaicing. A Long-Short Time Sliding Window (LST-SW) is designed to capture interested information during the mosaicing process.
\par
The rest of the paper is organized as follows: we reveal the mechanism of forming intra- and inter-frame artifacts in Section II. Section III elaborates the proposed blending method. In Section IV, we compare the experiment results and discuss the feasibility of the proposed method.  Concluding remarks and future work are given in section V.

\section{Background}
The degradation of sonar images during inspection tasks can be mainly classified into intra-frame artifacts and inter-frame artifacts. The causes and effects of both are explained below.

\subsection{Intra-frame artifact}
\begin{figure}[tb]
	\centering
	\subfloat[]{
		\includegraphics[width=1.6in]{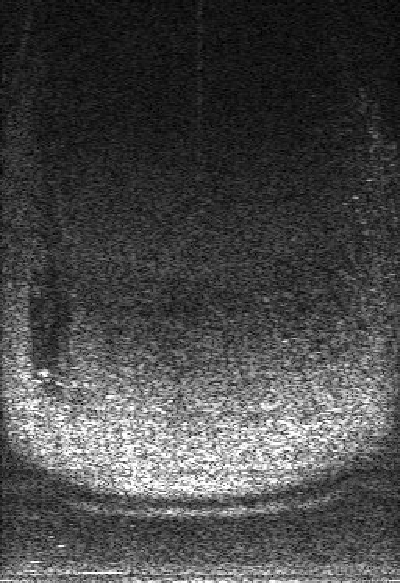}
		\label{intra_frame_artifact_bad}
	}
	% \hspace{-0in}
	\subfloat[]{
		\includegraphics[width=1.6in]{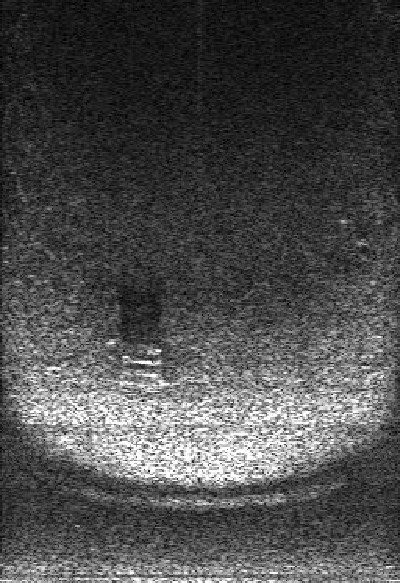}
		\label{intra_frame_artifact_good}
	}
	\caption{An open, box-shaped, lattice iron basket lying on the seabed imaged by FLS. (a) Strong reflection from the basket is overwhelmed by background. (b) A better result of another shot. The images are presented in raw (polar) format.}
	\label{intra_frame_artifact}
\end{figure}

The forward-looking sonar is recommended to work in a configuration that forms a small grazing angle between the boreline and imaged plane. This allows as large a volume as possible to be insonified. A smaller grazing angle results in an increase in the black area below the image (which has no reflected echoes), reducing the effective area of the image. Conversely, a greater angle exacerbates the effects of inhomogeneous insonification. As shown in Fig. \ref{intra_frame_artifact}, the strong intensity reflected by objects is overwhelmed by background reflection.

\begin{figure}[]
	\centerline{\includegraphics[width=3.5in]{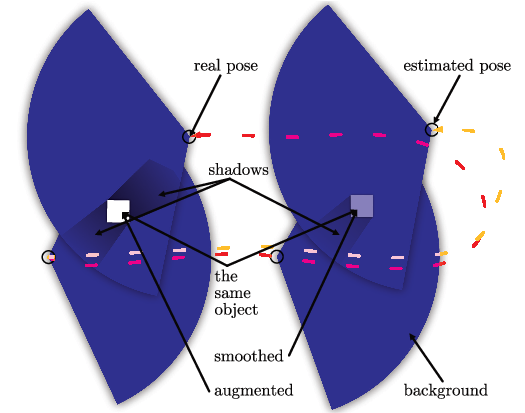}}
	\caption{Illustration of inter-frame artifacts. The brighter white square on the left side represents an angumentation. On the contrary, the fainter one on the right side represents a result of over-smooth.}
	\label{inter_frame_artifact}
\end{figure}

\subsection{Inter-frame artifact}
The inter-frame artifact is introduced by:
\begin{enumerate}
	\item object distortion caused by a change in viewpoint;
	\item inaccurate localization results.
\end{enumerate}
Object distortion can lead to a messy result, but not so far as to make objects totally fade away. However, the inaccuracy of localization may induce a misplacement of objects in the global coordinates of the mosaic, causing the object (interested information) to be smoothed by seafloor reverberation (useless information), as illustrated in Fig. \ref{inter_frame_artifact}.  The sonar is moving in lawn-mower pattern. The red dashed line indicates the real trajectory. The yellow dashed line indicates the estimated and error-accumulated trajectory, which introduces inter-frame artifacts. The object is smoothed by seabed background, instead of being augmented.
\par
The presence of intra- and inter-frame artifacts would severely impair the visual effect of the final mosaic. It further brings potential danger when the mosaic plays a role in object searching tasks for critical area. We explicitly take both of them into account and develop a method to preserve desired information from the artifacts during the mosaicing process.
\section{Proposed method}
The steps to generate FLS mosaic can be summarized as offset estimation and image blending. Here, we briefly describe the method adopted for offset estimation, then focus on the blending technology we developed. 

\subsection{Offset estimation}
During underwater inspection tasks, we get a sequence of single FLS images with different time stamps. In order to group them up in coordinate space, we should estimate the rotation and translation offsets between image pairs. FLS emits sound pulses and measures the reflected intensities to form an intensity map along beam and time directions. The insonified volume is a section of sphere. We denote a 3D point $\mathbf{p}$ in rectangular coordinates centered at sonar as $[x, y, z]^T$, with the $x$-axis looking forward. The point can also be represented in spherical coordinates as $\mathbf{s} = [r, \theta, \phi]^T$, where $r$ is the range distance between the sonar and the point, $\theta$ is the bearing angle, and $\phi$ is the elevation angle. The conversion between two representations can be achieved by the following equations:
\begin{align} % 公式：carte-polar互相转换
	\mathbf{p}&=\left[\begin{array}{c}
		x \\
		y \\
		z
	\end{array}\right]=\left[\begin{array}{c}
		r \cos \theta \cos \phi \\
		r \sin \theta \cos \phi \\ 
		r \sin \phi
	\end{array}\right], \label{equ:polar-carte-conversion} \\
	\mathbf{s}&=\left[\begin{array}{c}
		r \\
		\theta \\
		\phi
	\end{array}\right]=\left[\begin{array}{c}
		\sqrt{x^2+y^2+z^2} \\
		\arctan 2\left(y, x\right) \\
		\arctan 2\left(z, \sqrt{x^2+y^2}\right)
	\end{array}\right]. \label{equ:carte-polar-conversion}
\end{align}
Since every point at the same range along the elevation arc is projected into a single pixel, the elevation angle of the point is ambiguous. As a result, the imaging process of the FLS can be simplified from Eq. \ref{equ:polar-carte-conversion} as:
\begin{equation}
	\mathbf{x} =\left[\begin{array}{c}
		u \\
		v \\
	\end{array}\right]=\left[\begin{array}{c}
		r \cos \theta \\
		r \sin \theta
	\end{array}\right]. \label{equ:polar-carte-conversion-simplified}
\end{equation}
The projection can be approximated as an orthographic projection, and the offset between two sonar frames can be described as a combination of rotation and translation in sonar's moving plane. The pose of the sonar can be parameterized as $\mathbf{P}_t = [x_t, y_t, \theta_t]^T$, where $t \in \{1, 2, ..., N\}$ represents the number of the current frame, and $N$ is the total number of the sonar frames acquired in a task. We denote the transformation correlating the pixels in two consecutive frames as $\mathbf{T}_t \in \mathrm{SE}(2)$, and have\footnote[1]{The $\mathbf{x}$ is implicitly transformed between homogeneous coordinates and inhomogeneous coordinates}:
\begin{equation}
	\mathbf{x}_{t+1} = \mathbf{T}_t \mathbf{x}_{t}.
	\label{equ:xt-to-xt+1}
\end{equation}
We use the Fourier-Mellin Transform to estimate the $\mathbf{T}$, as done in \cite{hurtos2012fourier}. Firstly, the two images are  remapped in log-polar coordinates and then transformed to Fourier space in order to estimate the rotation angle. Then, the rotation is applied to one of the images. Finally, the translation between the rotated image and the other one is estimated through phase correlation. We refer readers to \cite{reddy1996fft}, for more details on the use of the Fourier-Mellin Transform in image registration.

\subsection{Blending method}
Now we  have images $\mathcal{I}$ and offsets $\mathcal{T}$ prepared. The most plainest strategy to generate a mosaic map $\mathcal{M} \in \mathbb{R}^{R\times C}$ is to average all the pixels at a certain global coordinate $(r, c)$, where $r \in R$ and $c \in C$, $R$ and $C$ are the height and width of the $\mathcal{M}$ respectively. The procedure is formulated as:
\begin{align}
	\mathcal{M}(r, c) &= \overline{\mathcal{M}_s(r, c)} \label{equ:avg1} \\
	&= \frac{1}{l(r, c)} \sum_{i=1}^{l(r, c)} p_{r, c, i}, \label{equ:avg2} \\
	\mathcal{M}_s &= \mathcal{F}(\mathcal{I}, \mathcal{T}), \label{equ:Ms}
\end{align}
where $l(r, c) = |\mathcal{M}_s(r, c)|$, $p_{r, c, i}$ is the $i$th pixel falling into $(r, c)$, and $\mathcal{F}$ is a function to locate each pixel from each image $I \in \mathcal{I}$ through $T \in \mathcal{T}$. We repeat this process until every $(r, c)$ pair is traversed. However, the presence of the intra- and inter-frame artifacts will definitely degrade the quality of $\mathcal{M}$.
\par
In \cite{hurtos2013novel}, the authors discovered that an inappropriate imaging configuration or significant relief variations can cause blind regions in the sonar frames, which is prone to smooth out useful image contents during the blending phase. We consider the reason behind that is, exactly, the disagreement in the image contents each pixel represents. We find, actually, the same reason causes the intra- and inter-frame artifacts.
\par
In addition to the decreased insonified regions, a too large grazing angle can also mask out the object lying on the imaged plane, as explained in Section II. The white pixels representing a strong or even saturated intensity can also serve a similar purpose as the blind regions do during mosaicing. As for the inter-frame artifact, the image content of a revisit frame with an error-prone pose estimation cannot precisely match the content of the original frame. In general speaking, the artifacts cause object pixels to be mixed with background pixels.
\par
With this in mind, we realize that simply averaging all the pixels that fall into a single coordinate is unfeasible. In the proposed method, we try to classify the pixels into $\mathcal{O}$ and $\mathcal{B}$, the union of object pixels and background pixels, respectively. At Equ. \ref{equ:avg1}, \ref{equ:avg2}, we only take $\mathcal{O}$ into the averaging process. This strategy isolates the negative impact of background pixels from the final map. We will then describe in detail how the classification is achieved.
\par
We utilize local second-order statistics $v$ for each pixel to discriminate the informative regions from the featureless seafloor. Then we apply the Long-Short Time Sliding Window (LST-SW) to $v$ to obtain score $s$. By stacking $s$ at the corresponding location, we get the Global Variance Map (GVM). Finally, we have $\mathcal{O}$ by GVM and then the mosaic.
\subsection{Local statistics}
Second-order statistics, commonly referred to as variance, is often used as a measure to indicate the data's dispersion. In \cite{hurtos2013novel}, the square root of the variance is employed to find the effective area of an FLS image. A lower value of $v$ indicates the flat seafloor containing no interested content. On the contrary, a higher value means that some informative regions are covered. We assume that all the FLS images have the same $K$ pixels in a task, and denote $v$ of a single pixel $p_{k, t}$ at timestamp $t$ as $v_{k, t}$, where $k \in \{1, 2, ..., K\}$, we get $v_{k, t}$ by:
\begin{align}
	\bar{p}_{k, t} &= \frac{1}{N_n}\sum_{i=1}^{N_n} p_{i, t}, \label{equ:mu} \\
	v_{k, t} &= \frac{1}{N_n}\sum_{i=1}^{N_n}(p_{i, t} - \bar{p}_{k, t})^2. \label{equ:v}
\end{align}
where $N_n$ is the number of pixels in the computation window centered at $p_{k, t}$.
\subsection{Long-short time sliding window (LST-SW)}
FLS images suffer from high-level speckle noise and inhomogeneous imaging patterns. The local statistics can also be affected. LST-SW is designed to counter those artifacts. The parameters of LST-SW consist of \(L_s\) and \(L_l\), which are the length of the sliding window for a short time and a long time, respectively. Literarily, $L_l$ is usually greater than $L_s$. The implementation of the LST-SW is formulated as follows:
\begin{align}
	v_{k,t,o} &= \sum_{t = -\frac{L_s}{2}}^{\frac{L_s}{2}} v_{k, t}, \label{equ:vo} \\
	v_{k,t,b} &= \sum_{t = -\frac{L_l}{2}}^{\frac{L_l}{2}} v_{k, t}, \label{equ:vb} \\
	s_{k,t} &= v_{k,t,o} e^{-v_{k,t,b}}. \label{equ:s}
\end{align}
\begin{figure}
	\centerline{\includegraphics[width=3.5in]{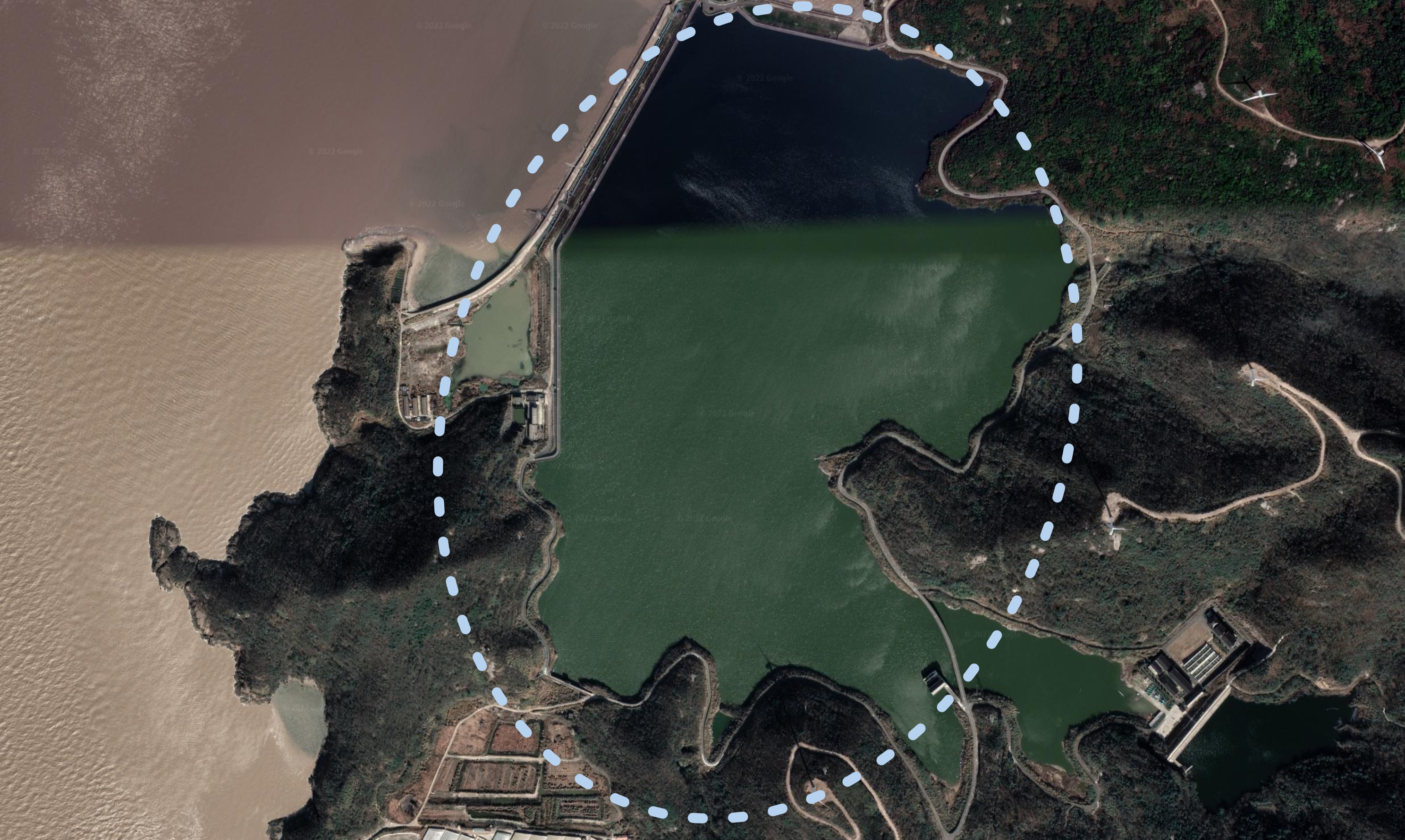}}
	\caption{The Huangjinwan reservoir which located on the northwest coast of Zhoushan City, Zhejiang Province, China. Image from Google Earth.}
	\label{field}
\end{figure}
where $s_{k,t}$ is the score representing the rectified local statistics for $p_{k, t}$. Since objects being imaged usually would be retained in several frames for a short time, averaging $v_{k, t}$ along the time-axis within \(L_s\) can reduce the value fluctuations, and furthermore, emphasize the salient area showing a consistently high value of $v$. However, the seabed and the illumination pattern of the sonar itself are more stable in time compared with the objects. Thus, averaging within \(L_l\) can model the statistics of background without the influence posed by objects. In Equ. \ref{equ:s}, we weight the $v_{k,t,o}$ by $v_{k,t,b}$ through an exponential mapping. We call this step Background Substraction. A $v_{k,t,o}$ accompanied with a high value of $v_{k,t,b}$ is more likely introduced by background. On the contrary, accompanying a low value of $v_{k,t,b}$ indicates a high probability of the presence of the interested objects. Finally, $s_{k, t}$ is obtained. By combining $s_{k, t}$ with the corresponding $p_{k, t}$ and the position in the global coordinates, we get a Global Variance Map (GVM).

\subsection{Global variance map (GVM)}
We denote GVM as $M_v$, which has the same dimension as $\mathcal{M}_s$ in the coordinate plane, but stores twice as many values at each position. We sort the $p$ in Equ. \ref{equ:avg2} in descending order by the corresponding $s$, and select the top $L_{\text{thres}}$ pixels into $\mathcal{O}$ for participating in the blending phase when forming the final mosaics. The use of GVM can avoid the contribution to the mosaic from the featureless seabed, which usually has lower $s$ values, competing with the valuable information.

\section{Experiments and results}
\begin{figure*}
	\centerline{\includegraphics[width=\textwidth]{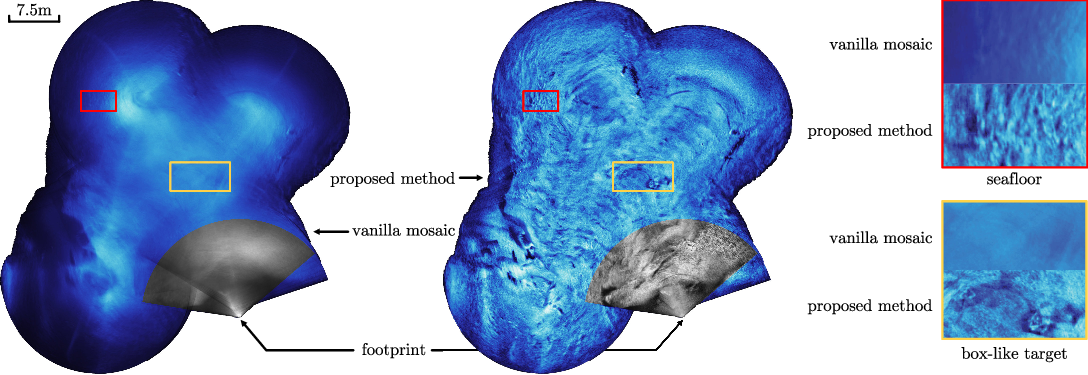}}
	\caption{The comparison of the mosaics constructed by the plain average and the proposed method. The zoomed areas are framed by different colors.}
	\label{mosaics_comp}
\end{figure*}
We have a set of sonar images captured by a Blueprint Subsea Oculus MD750d \cite{blueprintsubsea}. The operating parameters of the sonar are listed in Table \ref{tab:sonar-mode}. The sonar is mounted on an Autonomous Underwater Helicopter (AUH)\cite{chen2017computational}, a new type of autonomous underwater vehicle, and forms a small grazing angle with the bottom plane being observed. The AUH was about 3 meters above the bottom during the mission. The AUH performed a fold-back movement. The field of the experiment is the Huangjinwan reservoir as shown in Fig. \ref{field}. We collected a total of 3984 FLS images. During the computation of the algorithm, we only used half of the images making the computer memory affordable.
\begin{table}
	\caption{Working Mode of the Sonar}
	\label{tab:sonar-mode}
	\centering
	\begin{tabular}{|l|l|}
		\hline
		Name                       & Value  \\ \hline
		Operating frequency        & 1.2MHz \\ \hline
		Number of virtual beams    & 256    \\ \hline
		Number of Samples per beam & 373    \\ \hline
		Max range                  & 15m    \\ \hline
		Horizontal FOV             & 130°   \\ \hline
	\end{tabular}
\end{table}
\par
In this experiment, we set $L_s$ as 5, $L_l$ as 101, $L_{\text{thres}}$ as 15. The results are shown in Fig. \ref{mosaics_comp}. The gray sectors superposed on the mosaics are a sonar image that has been transformed to a cartesian coordinate. Table shows that the maximum detection distance is 15 meters, so the radius of the sector is 15 meters long. This gives an intuitive estimate of the size of the entire stitched image. The majority of the details are destroyed during the mosaicing process, as seen in the image on the left, leaving just the big mounds visible. Given the identical odometry estimates, the middle image produced by our approach displays a substantial amount of details. To further prove the superiority of our method, we manually choose two regions for zooming in and put them on the rightmost region of the Fig. \ref{mosaics_comp}. The two areas are a box-shaped target that we placed and the seafloor with a few rocks and sediments on it, respectively. It is clear that our method exposes more fine details. In particular, the plain average method causes the area to be over-smoothed due to the error accumulation in the offset estimates. While our method can work well since pixels with low $s$ value are excluded. Be aware that prior to image stitching, we processed each image using the Contrast Limited Adaptive Histogram Equalization (CLAHE) technique, which is identical to that in \cite{hurtos2013novel}. This technique somewhat removes the effects of brightness variation at different distances, but does not overlap with the role of our proposed method.

\section{Conclusion}
We analyze the intra- and inter-frame artifacts that could appear during underwater observation using FLS, and propose a new method suitable for image blending during the mosaicing of FLS images. The method utilizes the local statistics of sonar images, integrating LST-SW and GVM, to identify the pixels that are more likely to be rich in information as $\mathcal{O}$, so as to exclude the less informative pixels in $\mathcal{B}$ in final mosaic. An experiment using real sea-trial data shows that our method can provide more textural detail in the mosaic image and reveal originally invisible targets. This allows the operator to considerably reduce the inspection time after acquiring a large number of sonar images of an area, while not relying on automatic target recognition algorithms that are not yet well-developed. Concerning future work, we anticipate that the parameters can be adaptively tuned. We also hope that the algorithm will be able to run in an incremental manner to reduce memory and time consumption, and can be easily deployed on embedded systems for real-time execution.

\section*{Acknowledgements}
This work was supported in part by GDNRC[2021]32 and in part by the National Natural Science Foundation of China under Grant 62101489 and Grant 62171405. The authors would like to thank Bill Hanot at SoundMetrics Inc. for providing several datasets that played a crucial role in the preliminary validation work of the FLS image mosaicing algorithm. The authors would also like to thank Shaojian Yang for his review of the preliminary manuscripts of the paper.

\bibliographystyle{IEEEtran}
\bibliography{reference}

\vspace{12pt}

\end{document}